\documentclass[lettersize,journal]{IEEEtran}
\usepackage{amsmath,amsfonts}
\usepackage{algorithmic}
\usepackage{algorithm}
\usepackage{array}
\usepackage[caption=false,font=normalsize,labelfont=sf,textfont=sf]{subfig}
\usepackage{textcomp}
\usepackage{stfloats}
\usepackage{url}
\usepackage{verbatim}
\usepackage{graphicx}
\usepackage{xcolor}
\usepackage{cite}
\usepackage{soul,xcolor}
\sethlcolor{yellow}
\sethlcolor{orange}
\usepackage[hidelinks]{hyperref}
\usepackage{booktabs}
\usepackage{comment}

\begin{document}

\title{Estimation and Control of Tensegrity Manipulator Kinematics based on Strut Inclination Angles
}
\author{
    Tufail Ahmad Bhat$^{1}$,
    Shuhei Ikemoto$^{1}$
\thanks{Manuscript received: April 4, 2026; Revised: July 23, 2026; Accepted: September 11, 2026.}
\thanks{This paper was recommended for publication by Yong-Lae Park upon evaluation of the Associate Editor and Reviewers' comments.}
\thanks{This work was supported by the Japan Science and Technology Agency (JST) Fusion Oriented Research for Disruptive Science and Technology (FOREST) Program (Grant Number:  JPMJFR242H), the Japan Society for Promotion of Science (JSPS) KAKENHI (Grant Number: 25H02619), and the Japan Society for the Promotion of Science (JSPS) (Grant Number: 202621562).}
\thanks{$^{1}$All authors are with the Kyushu Institute of Technology, Japan (e-mail:  {\tt\small ikemoto@brain.kyutech.ac.jp})  }
\thanks{Digital Object Identifier (DOI): see top of this page.}
}

\markboth{IEEE Robotics and Automation Letters. Preprint Version. Accepted September, 2026}
{BHAT \MakeLowercase{\textit{et al.}}: TENSEGRITY ROBOT SHAPE ESTIMATION}

\maketitle

\begin{abstract}

Unlike conventional rigid-link robots defined by discrete joints, continuum robots pose a fundamental challenge for expressing their complex continuous bending configurations for closed-loop control. Several modelling approaches have been proposed for conventional continuum robots, but tensegrity-based continuum robots remain largely open. Moreover, many of these approaches assume a continuous elastic backbone and are therefore not directly applicable to tensegrity manipulators, whose bodies are networks of rigid struts and tensioned cables. 
This work presents a reduced-order model for shape and posture control of a tensegrity-based continuum manipulator. The manipulator is modelled as a serially connected parallel-link mechanism. The proposed method is formulated as an optimization problem that uses geometric constraints of the tensegrity structure together with information from the Inertial Measurement Unit (IMU) sensors embedded in the strut elements.  To the best of our knowledge, this work presents the first experimental demonstration of a real-time IMU-based shape estimation method on a full-scale tensegrity manipulator and demonstrates posture control using a simple Proportional-Integral (PI) controller. The results show that the proposed method can estimate the shape of both single-module tensegrity structures and multi-module tensegrity manipulators from arbitrary static configurations and achieve desired postures.
\end{abstract}


\section{Introduction}
 In rigid robotic systems, articulation is localized at the joints. At the same time, the links do not deform, so the entire shape, that is their configuration is uniquely determined by a set of joint variables. In contrast, soft and continuum robots lack discrete joints; their shape is therefore determined by a continuous distribution of deformation along their bodies. In theory, such robots possess infinite degrees of freedom because their configuration must describe deformation at every point along the body. As a result, there is no canonical low-dimensional choice of state variables that exactly captures their geometry \cite{bruder2020data}. To address this challenge, several approximation based modeling approaches have been proposed. Geometric-based methods rely on kinematic formulations, such as piecewise constant curvature (PCC) \cite{della2020improved} and polynomial curvature models \cite{della2019control}, which are computationally efficient and suitable for inverse kinematics and control. However, in their kinematic form they do not account for external forces, so the assumed curvature profile deviates from the true shape under external forces.  More accurate mechanics-based approaches, such as finite element methods (FEM) \cite{chaillou2023reduced} and Cosserat rod theory \cite{tummers2023cosserat}, explicitly account for external forces, but at higher computational cost. Sensor-based methods instead estimate shape directly from measurements. Shape sensing methods \cite{modes2020shape,wang2025shape}, measure local strain along the robot body using embedded sensors, from which curvature is inferred. For example, fibre Bragg grating (FBG) sensors. External shape sensing approaches \cite{shentu2023moss, zheng2024vision  } use vision-based sensors, but heavily rely on environmental conditions and are vulnerable to occlusions.  In real continuum robots, purely model-based methods are often insufficient due to modeling uncertainties. As a result, observer-based shape estimation methods have been proposed \cite{chen2019model, zhang2025stochastic }. These approaches combine physical models with sensor measurements to compensate for modeling errors.

\begin{figure}[!t]
    \centering
    \includegraphics[width=\linewidth]{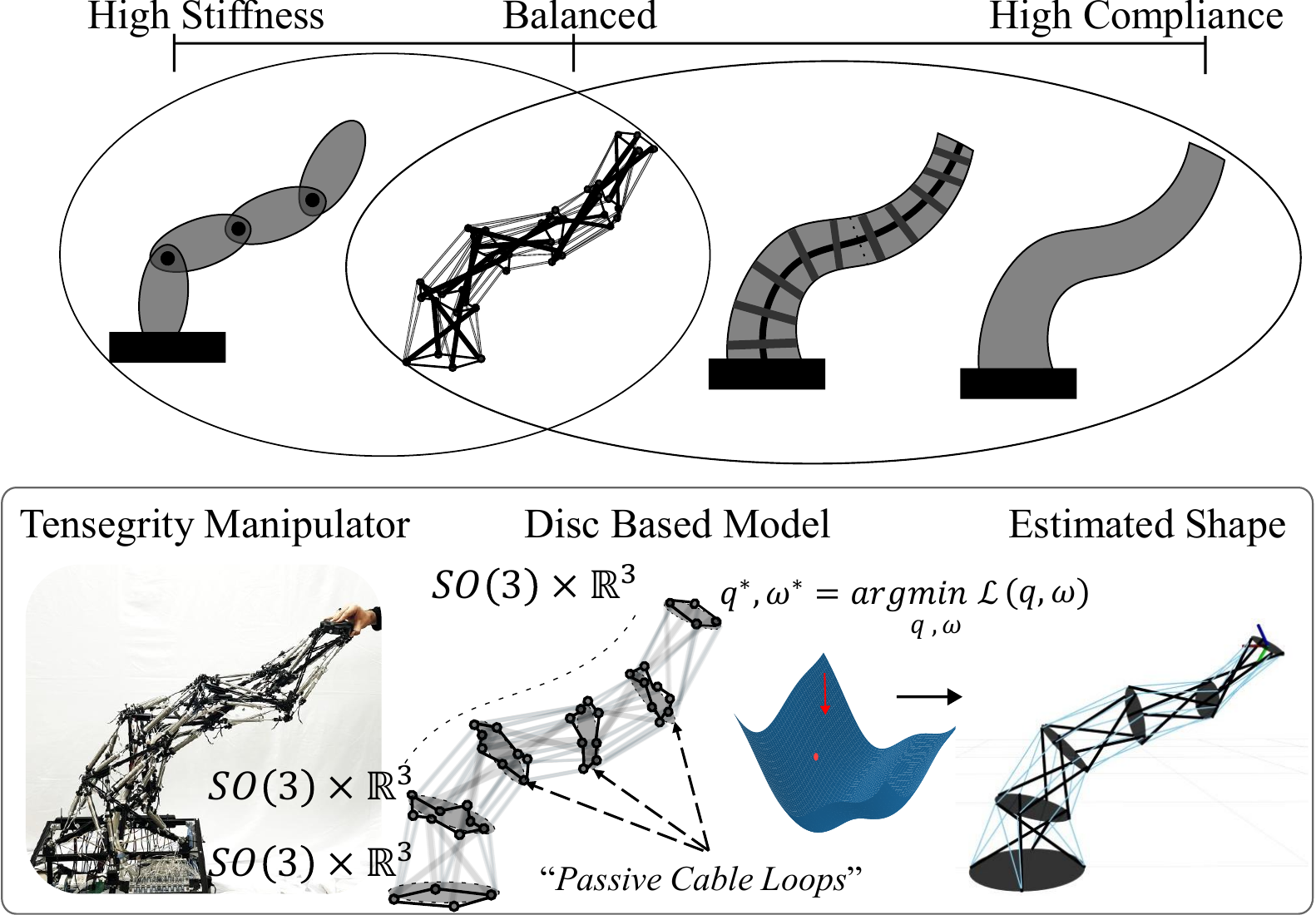}

    \caption{
   \textbf{Top}: A tensegrity manipulator lies between rigid robots and soft robots. \textbf{Bottom}: The TM-40 manipulator, a 20-strut tensegrity robot with 40 actuated cables and 40 passive cables. To estimate the overall shape, the manipulator is modeled by treating the passive cable as rigid discs that form a serially connected parallel-link mechanism. }
    \label{fig:TM40 tensegrity manipulator}
 
\end{figure}

Most of the aforementioned models assume an elastic backbone or a clearly defined continuous path, which simplifies the shape description to a curve along a single centerline. In contrast, tensegrity-based continuum manipulators have no such elastic backbone and instead rely on a network of cables and struts \cite{skelton2009tensegrity}. More fundamentally, the robot's shape is determined by force equilibrium in cables and struts rather than by curvature along the backbone. Tensegrity structures are lightweight, inherently compliant and can absorb large impact loads \cite{green2025steerable, cimatti2026modeling, johnson2026impact }, which has led to the development of modular tensegrity arms \cite{ramadoss2022hedra, lessard2016bio, fadeyev2019generalized,kobayashi2023active,10669185,kobayashi2022soft, 11246578,hsieh2024adaptive }, but knowing the shape remains a challenge for closed-loop control.  For simple single-module tensegrity robots, shape/state estimation has been approached using onboard cable-length sensors with external vision \cite{granados2026state}, or using internal strain sensors \cite{booth2021surface, mao2024multimodal, johnson2022sensor}. These sensors can provide accurate measurements, but they are often difficult to integrate, and scaling them to multimodule systems remains challenging. An alternative sensing option is inertial measurement unit sensors (IMUs), which are cost-effective and easy to integrate. Previous research has shown that IMU sensors are promising for reconstructing the posture \cite{santaera2015low}, and the kinematic model \cite{sato2024robot } of rigid robots. In deformable manipulators, previous studies \cite{stella2023soft, pei2025polynomial, cheng2022orientation,martin2022proprioceptive, pei2024imu } have also demonstrated the use of IMUs for shape reconstruction. In tensegrity-based robots, significant progress has been made, for instance, Wada \textit{et al.} \cite{10974957} utilised IMUs to define the posture of the tensegrity manipulator based on the tilt of the strut elements. Other studies \cite{caluwaerts2016state, bezawada2022shape, tong2025tensegrity} combined IMU sensing with external sensors to estimate the robot's state and shape. More recently, \cite{10955232} proposes an energy minimisation method over nodal positions that uses strut inclination information to estimate the shape of a single-module tensegrity structure, and their subsequent work \cite{bhat2026proprioceptive} extends to multimodule systems. However, it remains offline and shows increasing estimation errors under larger bending.

\begin{figure}[tb]
    \centering
      \includegraphics[width=\linewidth]{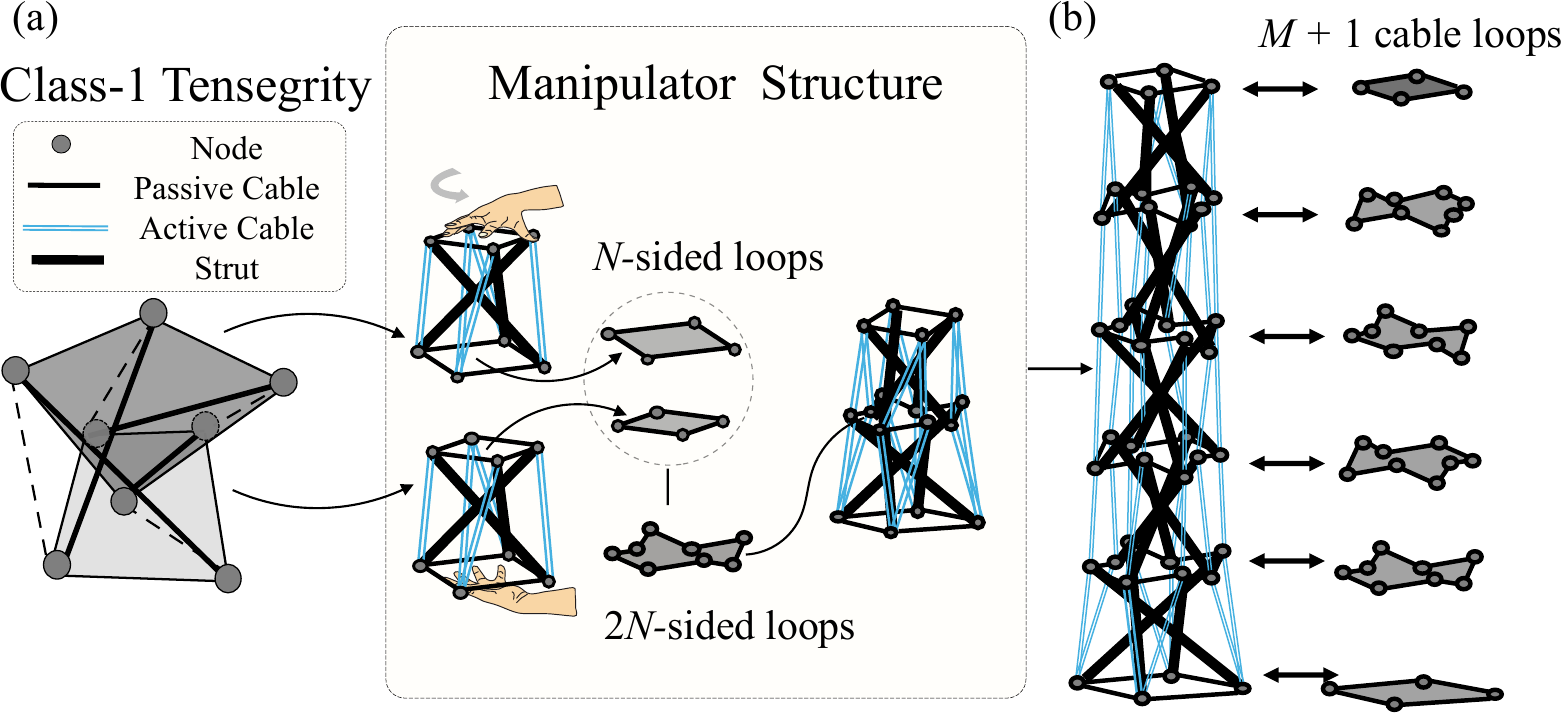}
    \caption{
   Schematic of the Tensegrity structure: a) A single-layer Class-1 tensegrity structure composed of 12 pre-tensioned cables and 4 rigid struts. b) The overall shape of the manipulator is formed by sequentially stacking the $M$ modules. This stacking creates $M+1$ highly tensioned passive cable loops, which represent the global shape of the structure.}
    \label{fig:stacking}
 \end{figure}

Most existing approaches focus on single-module tensegrity systems, and their scalability to a full-length tensegrity continuum manipulator with real-time performance and closed-loop control has not yet been demonstrated. In this work, we present an optimisation-based kinematic approximator for a modular tensegrity manipulator, which depends only on a single type of sensor, IMUs. Our main contributions are:

\begin{itemize}
    \item We propose an optimisation-based reduced-order model for shape and control of a tensegrity continuum manipulator, which only uses IMUs. 

 \item We validate the proposed method on a single-layer and multi-layer tensegrity robot by solving the shape estimation problem using a gradient descent (GD) method
  \item We demonstrate posture control for the tensegrity manipulator using a PI (Proportional Integral) controller based on the difference of the active cable lengths.
\end{itemize}
The proposed method uses the tensegrity structure's geometry as a constraint and approximates the passive cable as a serially connected parallel-link mechanism. It estimates the virtual pose of the moving disks that satisfy these constraints and IMU measurements from each strut. The validity of the method has been verified using a four-strut tensegrity prism and a five-layer tensegrity continuum manipulator shown in Fig.~\ref{fig:stacking}. In addition, we also demonstrate a simple PI controller based on the difference of the active cable lengths.
The structure of the paper is as follows. The proposed method is described in \hyperref[sec:]{Section II}. The details of the experimental setups, implementation, and experimental results appear in \hyperref[sec:experimental]{Section III}. Finally, the conclusions and directions for future work are summarised in \hyperref[sec:conclusion]{Section IV}.

\section{Proposed method}

\subsection{Class-1 Tensegrity Structure}
A tensegrity structure is made of compressive rigid elements, called struts, which carry compressive loads, and tensile elements, known as cables or strings \cite{skelton2009tensegrity}. The whole structure is stabilised by the pre-tension elements, and the overall configuration depends on the orientations of the rigid struts. Tensegrity structures are classified as class-\textit{k} systems, where up to \textit{k} rigid bodies are in contact. If \textit{k} = 1, the tensegrity structure becomes class-1, where no two rigid bodies are in contact, as shown in Fig.~\ref{fig:stacking} (a). Instead, the struts are suspended and stabilised by the tensile cable network. Among the different classes, "class-1 tensegrity" structures are pure tensegrity with no contact constraints, which makes the optimisation problem simpler.

\subsection{Approach}
The tensegrity manipulator consists of $M$ modules stacked sequentially to form a continuum structure, shown in Fig.~\ref{fig:stacking}. Each module contains $N$ struts. Two types of pre-tension elements are present. The first type consists of passive cables that form $M+1$ closed loop paths under high tension. These loops provide structural stability and can be reasonably approximated as rigid discs, which form the main idea of this work.  The basic assumption of the method is that each passive cable loop stays planar and approximately circular under high pre-tension. The second type consists of active cables (or actuators), placed between adjacent virtual discs, and changing the lengths of these cables produces bending of the manipulator. The proposed approach approximates the $M+1$ passive cable loops as discs or rings, as shown in Fig.~\ref{fig:stacking}, which together represent the overall shape of the manipulator.

\subsection{Formulation}
\begin{figure}[tb]
    \centering
      \includegraphics[width=\linewidth]{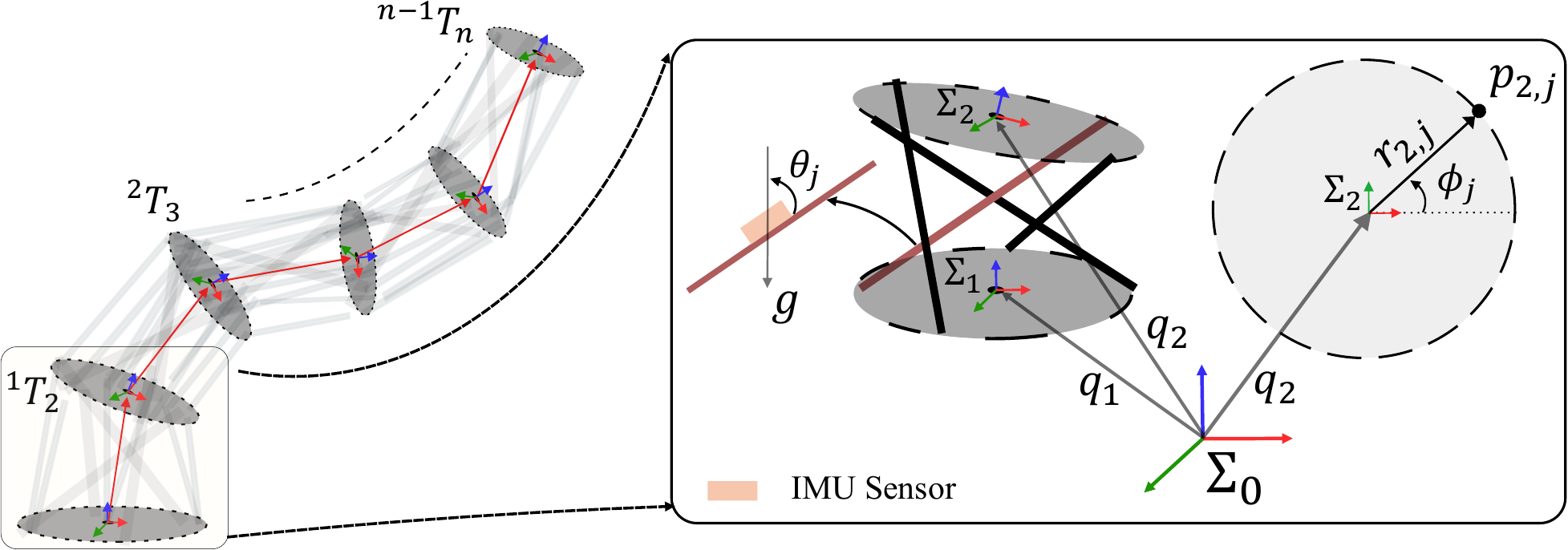}

    \caption{
   The figure shows the schematic of the bottom module of the manipulator,  the pose of the upper disk depends on the pose of the bottom disk, the strut length and the inclination of the strut element, which are obtained from IMU sensors equipped on each strut element. 
    }
    \label{fig:class 1}
 
\end{figure}

Consider a rigid circular disk with ball joints numbered from $0$ to $N-1$, evenly distributed along its circumference, such that the circle is divided into $N$ equal segments. Now consider two such disks. The ball joints with the same index on the two disks are connected by links $l_j$, where $j=0 \cdots N-1$, each of fixed length $D$. Let $\Sigma_1$ and $\Sigma_2$ denote the local coordinate frames attached to the centers of disks 1 and 2, respectively, shown in  Fig.~\ref{fig:class 1}. The position vectors, $q_i \in \mathbb{R}^3$, rotation matrices $R_i \in \mathrm{SO}(3)$, and rotation vectors (exponential coordinates) $\omega_i \in \mathbb{R}^3$ $(i=1,2)$ describe the pose of each disk with respect to the world coordinate frame $\Sigma_0$.
Furthermore, let $\theta_j$ denote the angle between link $l_j$ and the $z$-axis of the world coordinate system.

The problem can now be defined as estimating the most plausible position $q_2$ and orientation $\omega_2$ of the second disk, given that the pose ($q_1$, $\omega_1$) of the first disk, inclination angle and link lengths of each strut are known.
Assuming the radius of the rigid disk $i$ is $r_i$, the positions $p_{i,j}$ of the $N$ spherical joints on disk $i$ can be expressed:
\begin{eqnarray}
    p_{i,j} &=& q_i + \exp([\omega_i]_\times) r_{i,j}
\label{eq:disc_positions}
\end{eqnarray}
\begin{eqnarray}
    r_{i,j} = (r_i \cos \phi_j, r_i \sin \phi_j, 0)^T , \quad     \phi_j = \frac{2 \pi j}{N}
\end{eqnarray}

The length $D$ of the link $l_j$ and the inclination angle $\theta_j$ obtained from IMU sensor can be expressed based on $p_{i,j}$ using the following equations:

\begin{eqnarray}
  || p_{2,j}-p_{1,j} || =D, \quad 
    \cos \theta_j = \frac{(p_{2,j}-p_{1,j})^T e_z}{D} 
\end{eqnarray}

where $ e_z = (0,0,1)^T$ denotes the unit vector along the world $z$-axis

To estimate the relative pose of disk 2, we estimate the translation $q_2$ and rotation vector $\omega_2$ that satisfy the geometric constraints imposed by the structure. The most plausible $q_2$ and $\omega_2$ are obtained by solving the least-squares optimisation problem.

\begin{eqnarray}
    q_2^*, \omega_2^* = \arg\min_{q_2, \omega_2}\; \mathcal{L}(q_2, \omega_2)  
\end{eqnarray}

where the loss function $\mathcal{L}$ accumulates the squared residuals across all $N$ points,  and it can also be written as.

\begin{equation}
\mathcal{L}(q_2, \omega_2)=\frac{1}{2} \sum_{j=0}^{N-1}\left(\lambda_l\left[||v_j|| - D\right]^2+\lambda_z\left[ v_j^T e_z - D \cos \theta_j \right]^2\right)
\end{equation}

where $v_j = p_{2,j} - p_{1,j}$, and $\lambda_l$ and $\lambda_z$ are weighting coefficients . 

\begin{figure}[tb]
    \centering
      \includegraphics[width=\linewidth]{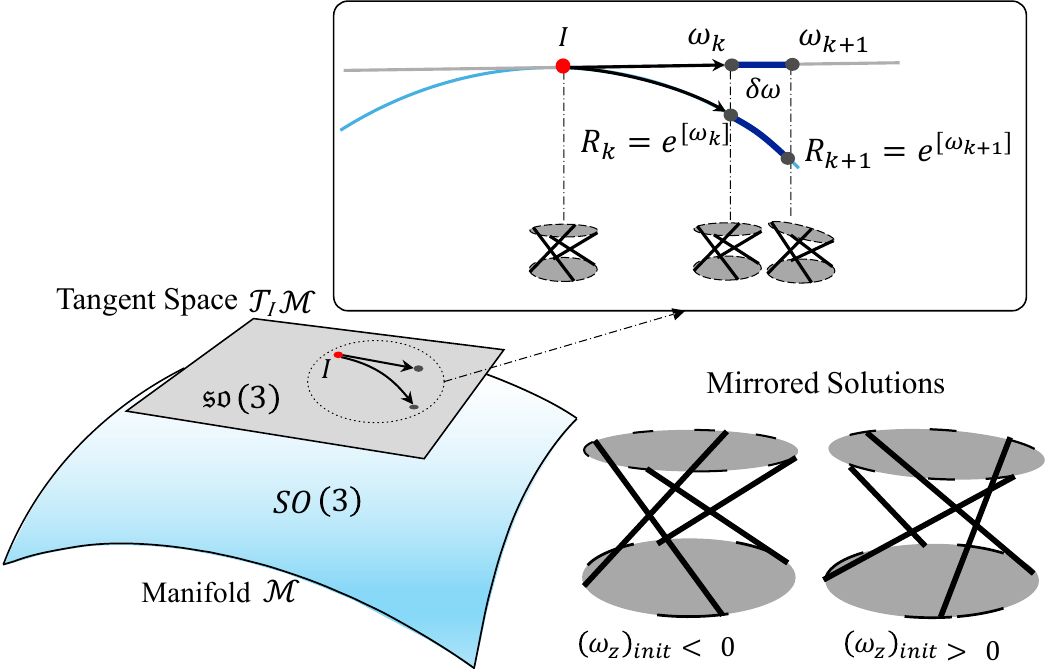}

    \caption{ The additive update on the tangent space at the identity of $\mathrm{SO}(3)$, followed by the exponential map as a retraction to obtain the rotation matrix of the discs. The bottom right side figure shows a mirrored solution. When an initial value of $\omega_z$ has the opposite sign, the solution can converge to a configuration with reverse twist direction. }
    \label{fig:manifold}
 \end{figure}

The optimisation problem is defined in a six-dimensional space, since the cost function $\mathcal{L}(q_2, \omega_2)$ depends on three-dimensional translation and rotation parameters. The rotation vector $\omega \in \mathbb{R}^3$ does not directly provide a linear operator that can rotate the points of the disc. To obtain a valid coordinate transformation, we convert the rotation vector $\omega = (\omega_x,\ \omega_y,\ \omega_z)$ using the hat operator and then exponential mapping. Specifically, the hat operator maps the rotation vector $\omega_2 \in \mathbb{R}^3
\;\overset{\widehat{\ }}{\longrightarrow}\;
[\omega_2]_\times \in \mathfrak{so}(3) $ to a skew-symmetric matrix
\[
[\omega]_\times
=
\begin{bmatrix}
0        & -\omega_z & \ \omega_y \\
\omega_z & 0         & -\omega_x \\
-\omega_y& \omega_x  & 0
\end{bmatrix}
\]
which lives in $ \mathfrak{so}(3) $ space, also known as \textit{Lie algebra}. And then the exponential mapping takes the $[\omega_2]_\times$ and gives us a valid rotation matrix of the discs $
[\omega]_\times \in \mathfrak{so}(3)
\;\xrightarrow{\;\exp\;}\;
\exp\big([\omega]_\times\big) \in \mathrm{SO}(3)
$ , the given matrix satisfies the $R^T R = I,\;\;
\det(R)=1$ rotation matrix constraints.

 \subsection{Gradient Computations}
We compute the analytical gradients of $\mathcal{L}(q_2, \omega_2)$ with respect to optimisation variables. For the translation component $q_2$, the final gradient can be written as:

\begin{equation}
\frac{\partial \mathcal{L}}{\partial q_2} = \sum_{j=0}^{N-1} \left[ \lambda_l ( ||v_j|| - D)  \frac{v_j^T}{||v_j||}+ \lambda_z (v_j^T e_z - D \cos \theta_j)   e_z^T \right]
\label{eqn:Lfinal/q2}
\end{equation}

And for the rotation component $\omega_2$, the final gradient is
\begin{equation}
\begin{aligned}
\frac{\partial \mathcal{L}}{\partial w_2} = \sum_{j=0}^{N-1} \Bigg [ ( \lambda_l (||v_j|| - D) \frac{v_j^T}{||v_j||} + \lambda_z (v_j^T e_z - D \cos \theta_j)  e_z^T ) \Big.\\ 
 (-\,R_2 [r_{2,j}]_\times J_{r}(\omega_2))  \Bigg]
\end{aligned}
\label{eq:Lfinal/w2}
\end{equation}

where $J_{r}(\omega_2)$ is the right Jacobian of SO(3). The resulting analytical gradients given in the eqs.~\ref{eqn:Lfinal/q2} and~\ref{eq:Lfinal/w2} enable us to use the gradient-based optimisers to find the optimal parameters $q_2^*$, $\omega_2^*$

 \subsection{Rotation Parameterisation and Jacobian}

 The rotation of the disc is parameterised by an axis-angle vector $\omega_2 \in \mathbb{R}^3$. As shown in Fig.~\ref{fig:manifold}, after each update, the corresponding rotation matrix is obtained using the exponential map, and it can be written as $R_2=\exp([\omega_2]_\times)$, where  $ [\;\cdot\;]_{\times} $ is a skew-symmetric matrix. Using Rodrigues rotation formula, the exponential map has the closed-form expression 
 \begin{equation}
    \exp([\omega_2]_\times) = I + \frac{\sin ||\omega_2||}{||\omega_2||}[\omega_2]_\times + \frac{1-\cos ||\omega_2||}{||\omega_2||^2}[\omega_2]^2_\times
\end{equation}

 Differentiating the above expression gives us the right Jacobian of SO(3), which is used to compute the $\nabla_{\omega_2} \mathcal{L}$ in eqn.~\ref{eq:Lfinal/w2}. It can be expressed as follows:
\begin{equation}
    J_{r} = I - \frac{1-\cos ||\omega_2||}{||\omega_2||^2}[\omega_2]_\times + \frac{||\omega_2||-\sin ||\omega_2||}{||\omega_2||^3}[\omega_2]^2_\times
\end{equation}
 
During the optimisation, after each step, the rotation matrix is updated as $R_2 \leftarrow \exp\! \left([\omega_2]_\times\right)$. For small $||\omega_2||$, the higher-order terms in the series expansion vanish rapidly, and it is recommended to approximate it with an initial term as 
\begin{equation}
J_{r} \approx I-\frac{1}{2}[\omega_2]_\times + \frac{1}{6}[\omega_2]^2_\times
\end{equation}
The equation maintains numerical stability and ensures smooth convergence.
\begin{algorithm}[t]
\caption{Geometry Constrained Shape Estimation}
\label{alg:shape_estimation}
\begin{algorithmic}[1]
\STATE \textbf{Input:}  Measured angles $\theta_j$ obtained from IMUs, strut length $D$, disc radius $r$, number of struts $N$
  \STATE \textbf{Output:} Estimated pose $(q_2^*, \omega_2^*)$

\STATE \textbf{Initialize:} $q_2 $, $\omega_2$

\FOR{$k = 1$ to $K$}
    \STATE Compute joint positions: $p_{i,j} = q_i + R_i r_{i,j}$                           (\ref{eq:disc_positions}) 
    
    \STATE Compute residuals:
    \FOR{$j = 0$ to $N-1$}
        \STATE $v_j = p_{2,j} - p_{1,j}$
        \STATE $\ell_j \gets \|v_j\| - D$ \COMMENT{Length constraint}
        \STATE $a_j \gets v_j^T{e}_z - D\cos\theta_j$ \COMMENT{Angle constraint}
    \ENDFOR
    
    \STATE Compute gradient: $\mathbf{g} = \nabla_{q_2,\omega_2} \mathcal{L}$
    
    \STATE Update: $q_2 \gets q_2 - \eta_q \mathbf{g}_q$, $\omega_2 \gets \omega_2 - \eta_\omega \mathbf{g}_\omega$
    
    \STATE Update rotation: $R_2 \gets \exp([\omega_2]_\times)$ 
    \ENDFOR

\RETURN $q_2$, $\omega_2$
\end{algorithmic}
\end{algorithm}
 \subsection{M-layer Tensegrity Pose estimation}
 To extend the proposed method from two layers to $M$ modules, we use a sequential layered optimisation. The pose of each disc is estimated from the preceding disc's pose. Let $\mathcal {}D_i$ denote disc $i (i=2, \ldots, M+1)$, where $\mathcal {}D_1$ is fixed to the ground. The pose of disc $i$ is described by a translation vector $q_i \in \mathbb{R}^3$ and a rotation matrix  $R_i \in \mathrm{SO}(3)$. The pose of a disc $i$ depends on the pose of disc $i-1$, and can be written as
 $$
\begin{aligned}
q_i & =q_{i-1}+R_{i-1} q_i^{\text {rel }} \\
R_i & =R_{i-1} R_i^{\text {rel }}
\end{aligned}
$$
where $q_i^{\text {rel }}$ and  $R_i^{\text {rel }}$ are the translation and rotation of disc $i$ expressed in the coordinate frame of disc $i -1$ . Accordingly, the optimisation variables for each layer are the 6 DOF relative pose parameters.

$$
\mathbf{T}_i=\left[q_i^{\text {rel }} | \omega_i^{\text {rel }}\right] \in \mathbb{R}^6 
$$

 In the implementation, each module is optimized independently using its own optimisation loop and hyperparameters. 

\section{Experimental Results}

This section provides the details of the experimental hardware configurations and evaluates the proposed Algorithm 1 on two tensegrity platforms.  We then present the corresponding estimation results under different conditions, and apply a simple PI controller for whole body control, and finally discuss the findings.

\begin{figure}[t]
 \centering
      \includegraphics[width=\linewidth]{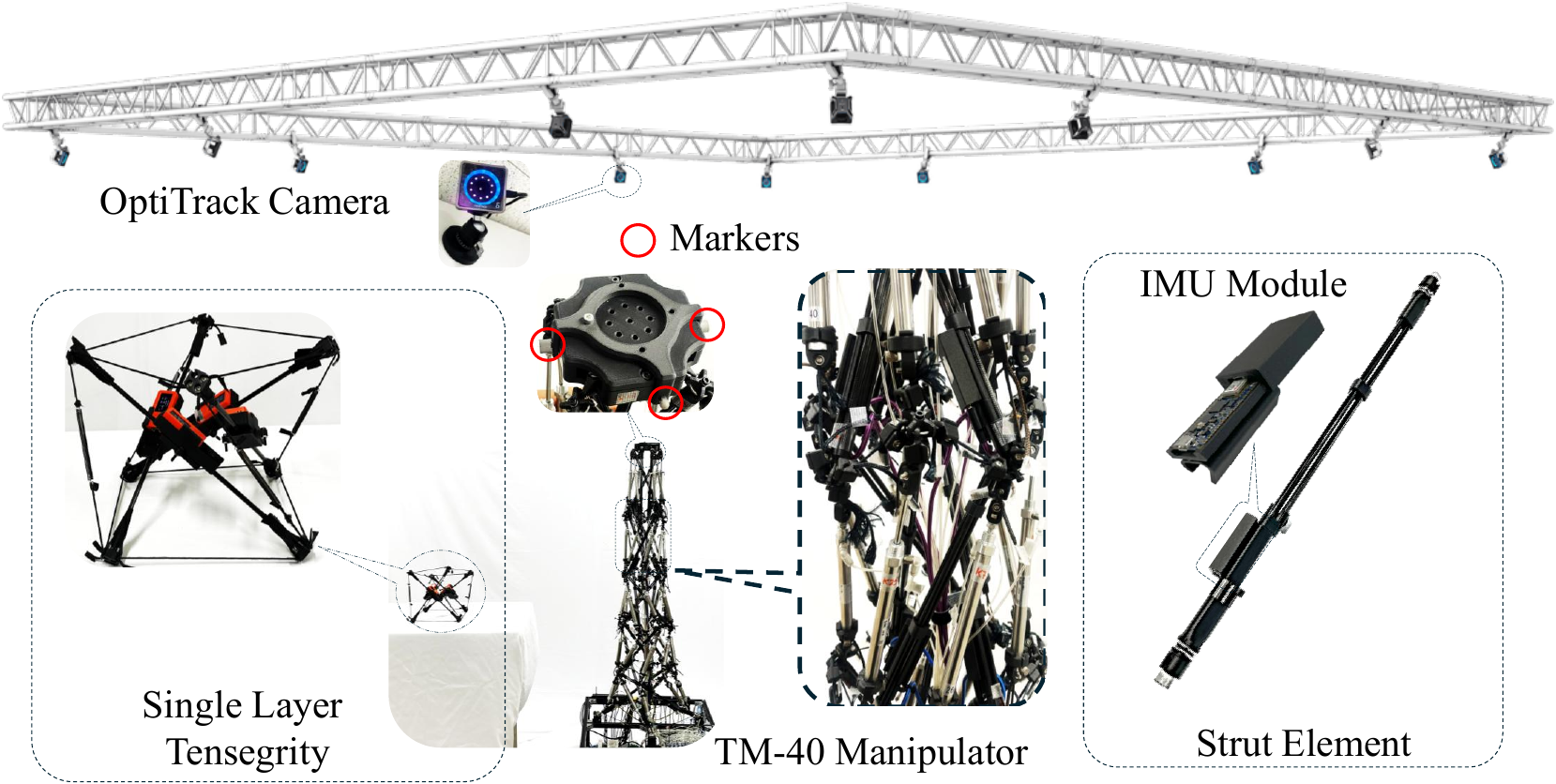}
     \caption{The motion capture experimental space showing a single-layer tensegrity structure and a modular 5-layer tensegrity manipulator, with IMU sensors attached to each strut that provide inputs to the shape estimator. }
     \label{fig:whole_setup}

\end{figure}
 \subsection{Experimental Setups}

 Figure~\ref{fig:whole_setup} shows the experimental environment, with the two setups placed side by side. The first setup is a single-layer class 1 tensegrity, which consists of four struts and twelve pre-tensioned cables. Each strut element is equipped with an M5StickC Plus 1.1 device. It contains an ESP32-PICO-D4 dual-core microcontroller, a built-in IMU (MPU6886), and a wifi module. The 6-axis IMU sensor provides inclination angle measurements with respect to global coordinates. The main setup is the 20-DOF TM40 manipulator, which consists of five class-1 tensegrity modules. Each module contains 4 struts and 8 active cables (pneumatic actuators). The cylinders are arranged vertically and grouped into 8 directions to enable bending control. Actuation is achieved by regulating the pressure inside the cylinder using 40 independent pressure control valves (VEAB, FESTO). The operating pressure range is 0.1 - 0.6 MPa. In total, the TM40 includes 20 struts, each equipped with an Arduino Nano RP2040 with an onboard IMU sensor (LSM6DSOX), which provides inclinational angle measurements. Three reflective markers are attached to the tip of the manipulator, and an OptiTrack motion capture system with 8 cameras tracks the tip pose
 to provide ground truth measurements.
 \begin{figure*}[t]
    \centering
     \includegraphics[width=\linewidth]{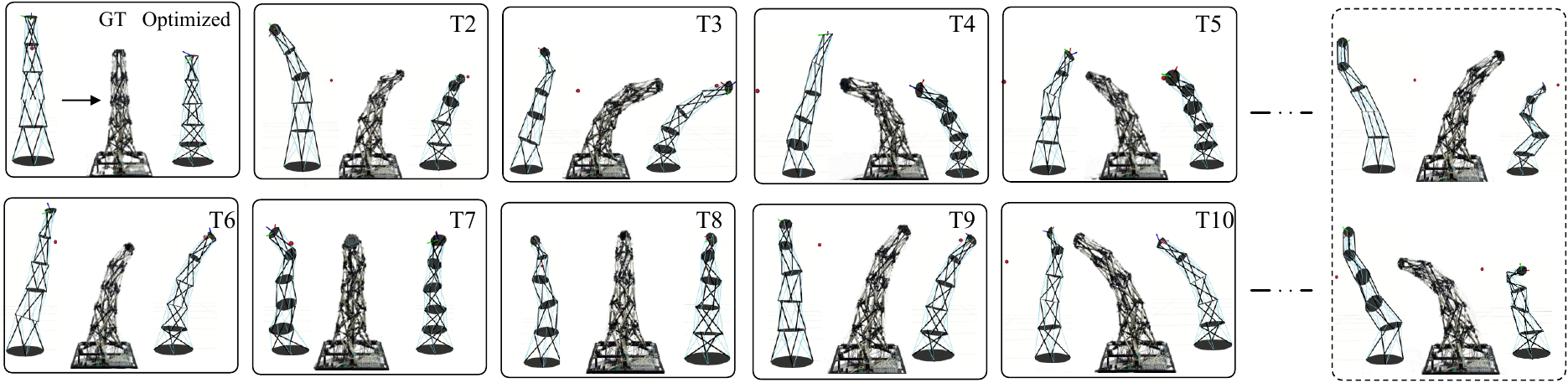}

    \caption{ The manipulator was actuated into 20 different static postures, with the estimator initialized randomly for each trial. \textbf{Left:} Successful cases: shows snapshots of the shape after the first optimisation step and then the converged shape. \textbf{Right:} Examples of failed solutions, where the estimator has not converged to a physically plausible shape.  (GT: Ground Truth)}
    \label{fig:arbitary_values_for_TM40}
 \end{figure*}
 \begin{figure}[tb]
    \centering
      \includegraphics[width=\linewidth]{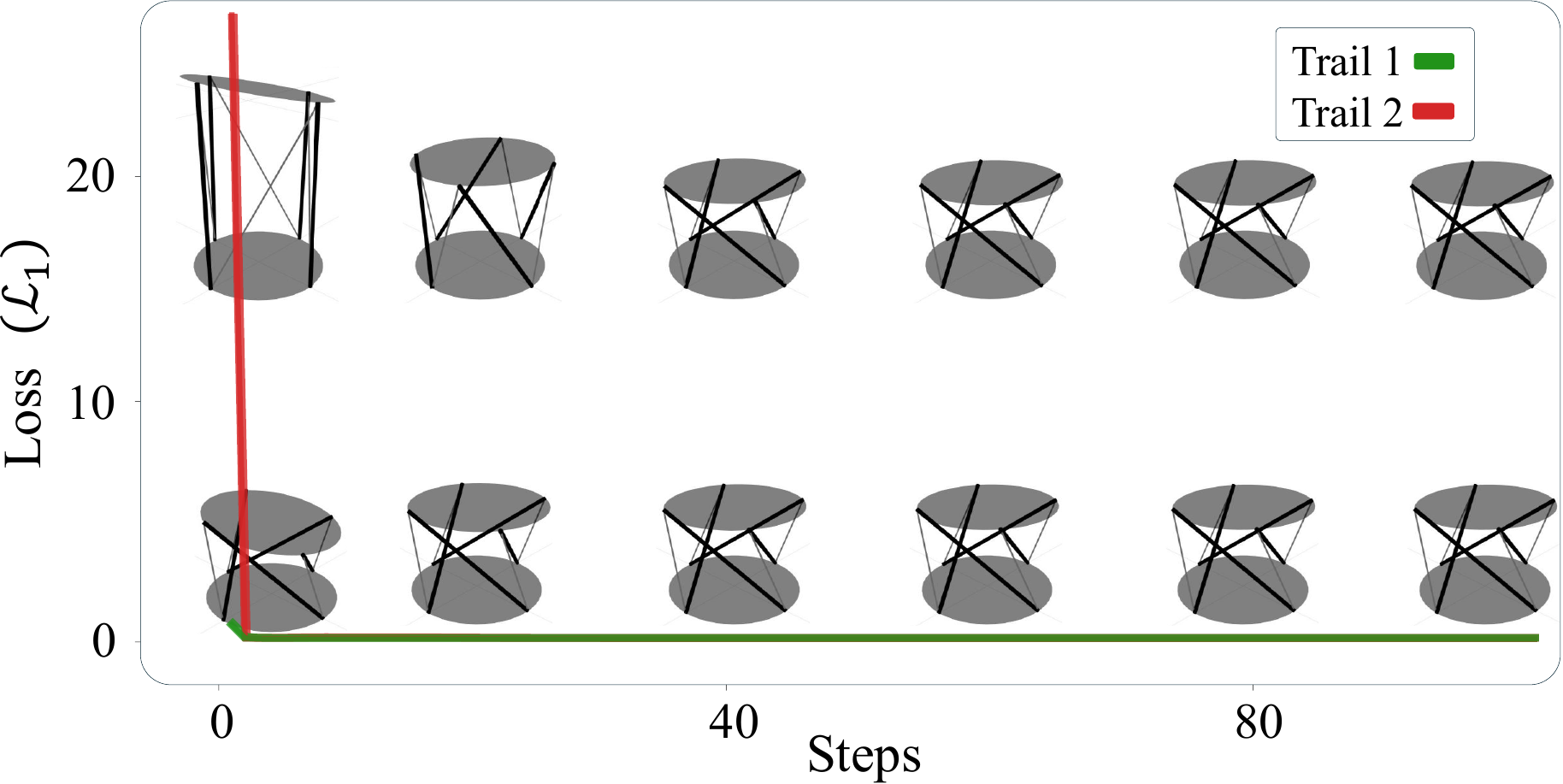
}
    \caption{ The estimator is initialized with expanded (red) and collapsed (green) initial shape estimates, and both trials show convergence.}
    \label{fig:single_layer_extreme}
 \end{figure}
\begin{figure}[tb]
    \centering
      \includegraphics[width=\linewidth]{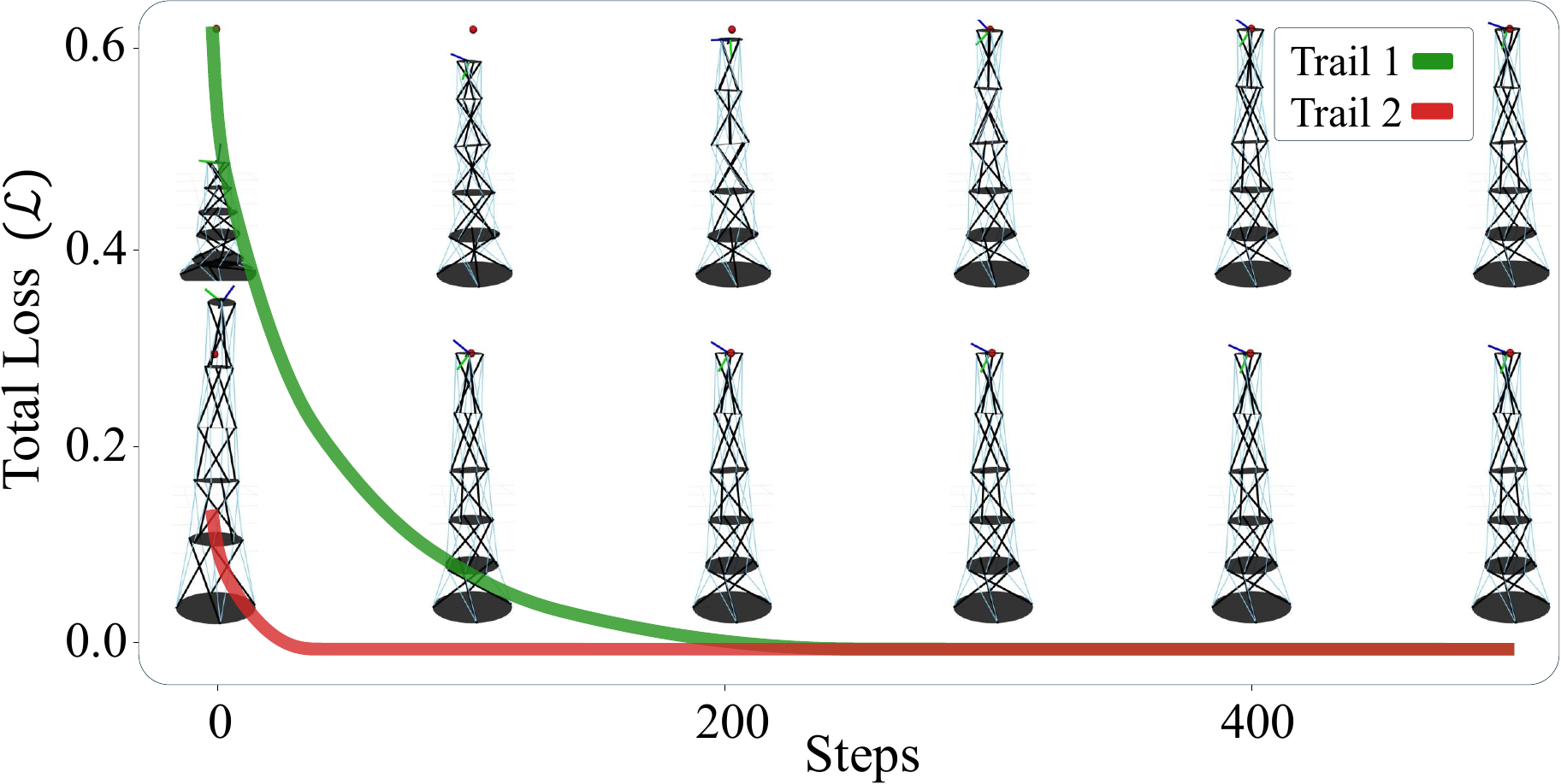}

    \caption{
  Decrement of the objective functions, both trials show the convergence of the optimisation problem}
    \label{fig:extreme_TM40}
 \end{figure}

\subsection{Shape Estimation Evaluation}

\subsubsection*{Extreme Initialization Convergence }
In this experiment, both experimental setups remain in a static configuration as shown in Fig.~\ref{fig:whole_setup}. Then the shape estimator is initialized with extreme values of $q$ and $ w$, corresponding to collapsed and expanded configurations, and the resulting objective is solved numerically using the GD method. For each setup, we conducted two trials. After the first optimisation step, the objective function values are 0.60 and 26.16 for trials 1 and 2, respectively. The Fig.~\ref{fig:single_layer_extreme} shows that the loss fluctuates around $9.58 \times 10^{-7}$ and $1.56 \times 10^{-6}$ units as the optimization steps proceeded to 100 steps, and the estimated shapes stabilized and aligned with the actual setup of single layer class-1 tensegrity.
 \begin{table}[h]
\centering
\caption{Shape estimator hyper parameters for TM40 manipulator}
\label{tab:hyperparams}
\begin{tabular}{l c}
\toprule
Parameter & Value \\
\midrule
Strut lengths $D_1\ldots D_5$ & 0.39, 0.32, 0.32, 0.32, 0.32 m \\
Disk radii $r_1\ldots r_6$ & 0.20, 0.135, 0.115, 0.10, 0.07, 0.066 m \\
 $\lambda_l$  & 1.0 \\
 $\lambda_z$  & 1.0 \\
$\eta_q$ & 0.13 – 0.18  \\
$\eta_w$ & 0.4 – 0.8  \\
GD iterations per update & 1 \\
\bottomrule
\end{tabular}
\end{table}

Similarly, for the TM40, the whole-body shape is reconstructed using sequential optimisation, where each layer is solved given the pose of the previous disc and the geometry constraints of the current disc. The hyperparameters used in the experiments are shown in Table~\ref{tab:hyperparams}. The total loss $\mathcal{L}$ is the summation of the individual module losses $\mathcal{L}_1$, $\mathcal{L}_2$, $\mathcal{L}_3$, $\mathcal{L}_4$,and $\mathcal{L}_5$. Figure~\ref{fig:extreme_TM40} shows a decrement in the objective function after approximately 250 optimisation steps, converging to a neighbourhood of $5.63 \times 10^{-5}$ and $5.48 \times 10^{-5}$ in trials 1 and 2, respectively.

 \begin{figure}[tb]
    \centering
    \includegraphics[width=\linewidth]{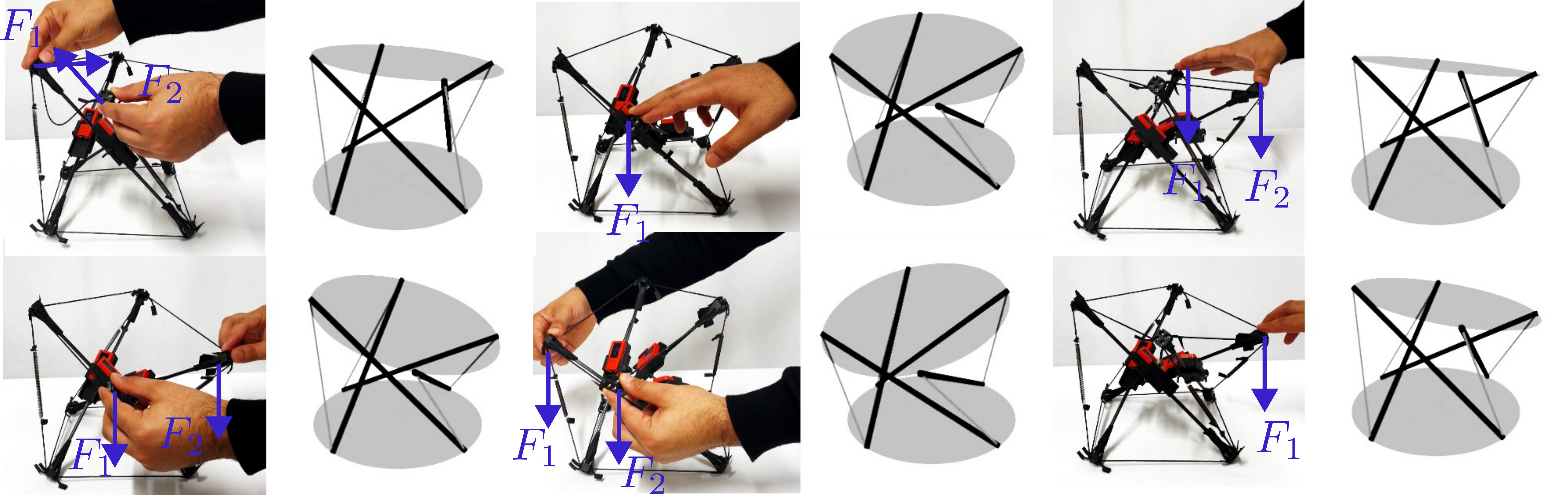}

    \caption{
  Under external manual deformations ($F_1$-$F_2$), the estimator runs online, and the estimated shape tracks the deformations. The sequence of postures of the single layer class 1 tensegrity structure and their corresponding shapes are visualized in RViz.}
    \label{fig:class1_dynamic}
 \end{figure}
 
\subsubsection {Arbitrary Configuration Shape Estimation}

 \begin{figure}[tb]
    \centering
  \includegraphics[width=\linewidth]{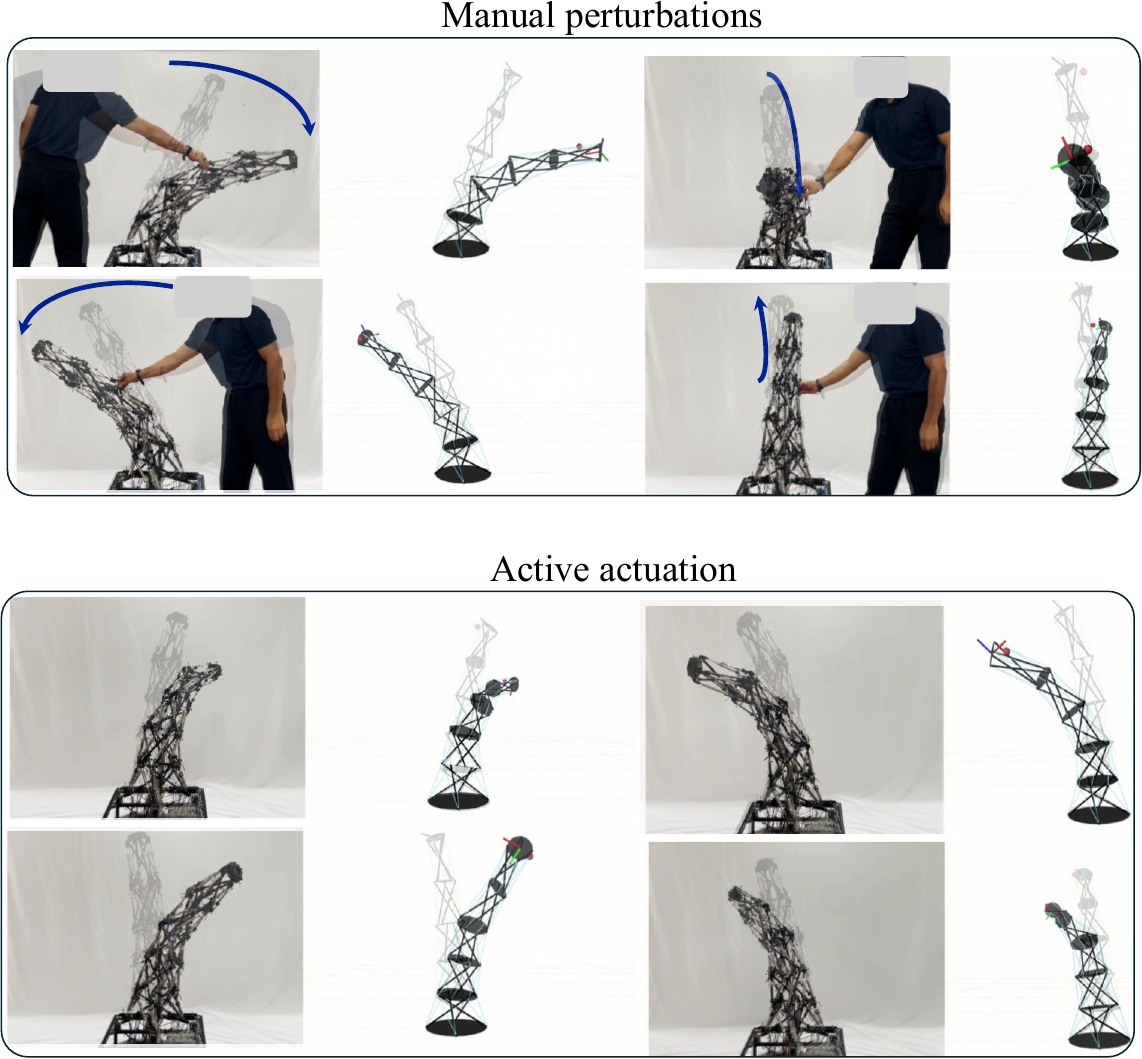}

    \caption{ Sequential shape estimation during \textbf{Top:} manually induced motion and \textbf{Bottom:} active actuation. The GD method runs online to solve the optimisation problem, the corresponding tip trajectory errors along each axis are shown in Figure~\ref{fig:spatial_positions}.}
    \label{fig:sequential_shape}
 \end{figure}
In this experiment, we set the pressure values so that the manipulator covers bending in all directions. We then initialise the unknown variables randomly, without prior knowledge of the true shape. The estimator
is then executed, as shown in Fig.~\ref{fig:arbitary_values_for_TM40}, in total, 20 experiments were tested. Our method converged to the corrected shape in 16 out of 20 cases. For each successful configuration, we evaluated both tip position and orientation using the (MoCap) system. For the tip position, the root mean square error (RMSE) across all trials
is approximately 0.0913 m, with a maximum error of 0.147 m. The tip orientation error is decomposed into tilt (roll and pitch) and yaw components, the RMSE tilt and yaw errors are approximately 0.216 and 0.205 radians, with maximum errors of 0.357 and 0.371 radians, respectively.
 \begin{table}[tb]
\caption{Positional and orientational errors of the tip of the robot after convergence}
\label{tab:ee}
\centering
\footnotesize
\setlength{\tabcolsep}{4pt}
\begin{tabular}{cccccc}
\toprule
Trial & Pos. error (m) & Tilt error ($rad$) & yaw error ($rad$) \\
\midrule

T1 & 0.026 & 0.061 & 0.020 \\
T2 & 0.147  & 0.357 & 0.268 \\
T3 & 0.124 &  0.149& 0.122\\
T4 & 0.115  & 0.124 & 0.298 \\
T5 & 0.067 & 0.267 & 0.205 \\
T6 & 0.068 & 0.197 & 0.019 \\
T7 & 0.067 & 0.111 & 0.052 \\
T8 & 0.061 &0.057& 0.078 \\
T9 & 0.039 &0.290& 0.371 \\
T10 & 0.117 & 0.298 & 0.235 \\

\midrule
\multicolumn{4}{c}{Position RMSE 0.0913 \,m\ - Max 0.147  \,m} \\
\multicolumn{4}{c}{RMSE tilt $0.216$, max $0.357 $ rads : RMSE yaw $0.205$, max $ 0.371$ rads } \\
\bottomrule
\end{tabular}
\end{table}
Table~\ref{tab:ee} reports the corresponding errors across all the trials. In addition, the failed cases can be discarded based on the estimated disc poses violating the geometric constraints of the structure. Once the solution has converged, the estimator can be used to track the robot's shape dynamically. For the single-layer case shown in Fig.~\ref{fig:class1_dynamic}, the estimator follows the shape during manual deformation.

\subsection{Dynamic Motion Evaluation}
\begin{figure}[t]
    \centering
  \includegraphics[width=\linewidth]{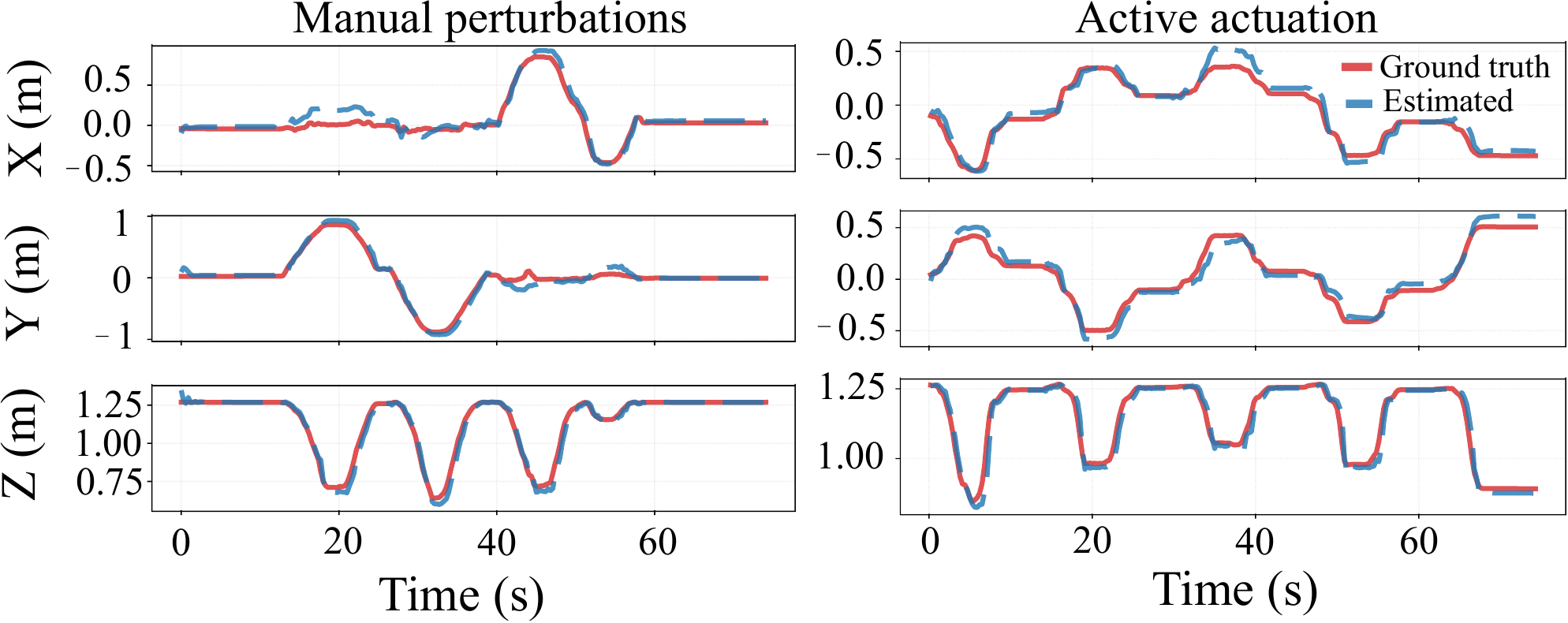}

    \caption{Spatial position of the tip of the manipulator: \textbf{Left:} Tip position of the TM40 manipulator under \textbf{Left:} manual pertubation and \textbf{Right:} active actuation.  Estimated trajectories (blue) are shown against MoCap ground truth (red) for each axis. }  
    \label{fig:spatial_positions}
 \end{figure}

To evaluate the online performance of Algorithm 1, we conducted two dynamic experiments. In the first experiment, a human manually perturbed the robot in all directions. In the second experiment, the manipulator bent continuously under its own pneumatic actuation.  In each case, the estimated tip position ($q_6^*$) is compared with the ground-truth measurements from the MoCap. The RMSE over the whole trial was 0.1024 m for manual perturbation, and 0.0990 m for the active case.  Although multiple sensors introduce latency, the estimated tip trajectory approximately follows the ground truth measurements and shows stable convergence, as shown in Fig.~\ref{fig:sequential_shape}. The corresponding tip position trajectories are shown in Fig.~\ref {fig:spatial_positions}.

\begin{figure}[t]
    \centering
  \includegraphics[width=\linewidth]{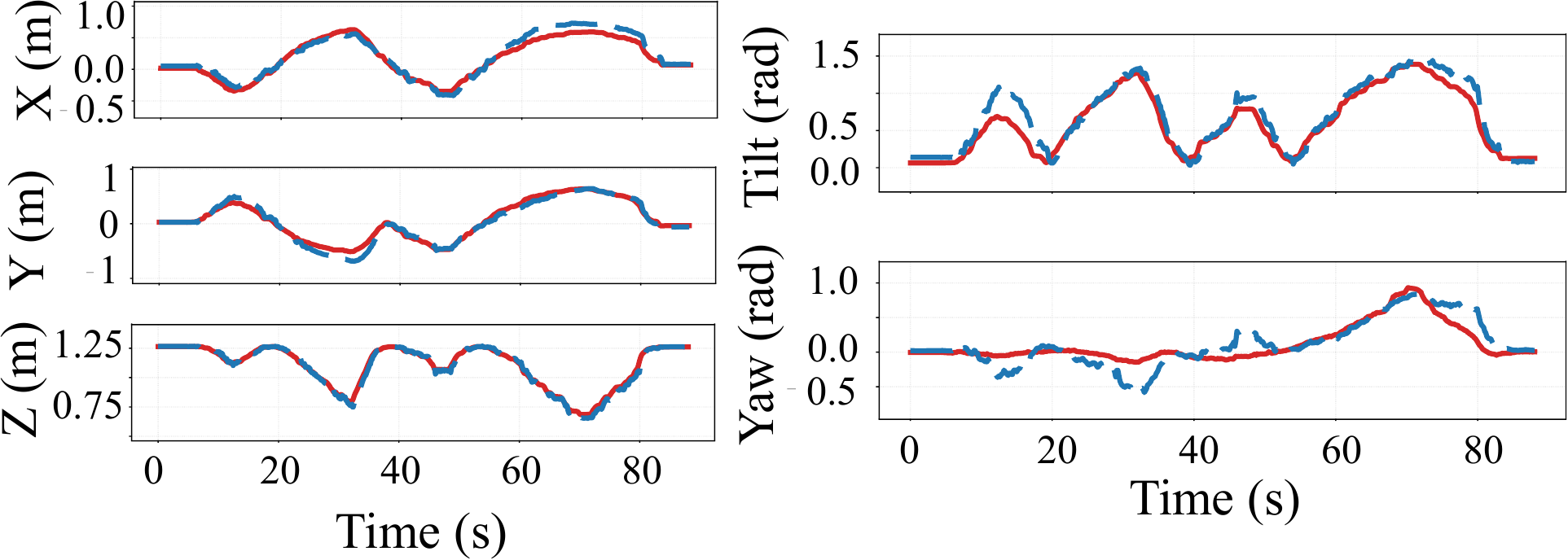}

    \caption{  Tip pose during dynamic tracking. \textbf{Left:} Tip positions for each axis. \textbf {Right:} Tilt and yaw angles over the same interval. } 
    \label{fig:pose_errors}
 \end{figure}

 To further evaluate the full 6-DOF tip pose, we conducted an additional manual perturbation experiment while simultaneously recording the MoCap tip pose. Figure~\ref{fig:pose_errors} shows the 3D tip positions, corresponding tilt (roll/pitch) and yaw angles over an 80s interval. After an initial calibration, the RMSE for tilt and yaw angles is approximately 0.144 and 0.171 radians, respectively, and the RMSE for 3D tip position is around 0.113 m, which is approximately consistent with the above reported error in the manual-perturbation case. In the dynamic case, the orientation errors are lower than the static ones in Table~\ref{tab:ee}, this may be because the estimator starts each step from the previously converged solution.

 \subsection{Computation analysis}
 
The proposed method was implemented in Python 3.10, and the experiments were performed on a desktop computer with an Intel Core i9-10885H processor and Ubuntu 22.04 LTS as the operating system, with ROS 2 Humble as the software environment. In addition, the estimated shape is visualized in ROS 2 using RViz 2 via \texttt{visualization\_msgs/Marker} messages.

\begin{table}[h ]
\centering
\caption{Computational time of the estimator}
\label{tab:Computational_time}
\begin{tabular}{ccccc}
\toprule
Run & Avg Time/Step (\textit{ms}) & Std (\textit{ms}) & Min (\textit{ms}) & Max (\textit{ms}) \\
\midrule
1 & 3.12 & 1.10 & 1.90 & 10.46 \\
2 & 2.49 & 0.87 & 1.89 & 10.54 \\
\bottomrule
\end{tabular}
\end{table}

We measure the computation time per optimization iteration, where one iteration corresponds to a single GD update. Each module consists of four IMUs, and in total 20 IMU sensors. The IMUs publish measurements at approximately 80 Hz, and the data are received asynchronously by a ROS 2 subscriber. In an initial implementation, a single ROS~2 node processed all 20 IMU streams, with each incoming IMU measurement triggering an optimisation loop independently. On average, a single optimisation loop takes 2-3 \textit{ms}, but the callback queue grows without bound, since the 20 streams trigger approximately 1600 optimizations per second. To reduce the computation load, we run the estimator at a fixed rate. A dedicated timer then reads the latest measurements and performs a single GD step across all five layers at 80 Hz rate. Table~\ref{tab:Computational_time} shows the runtime performance of our proposed method.

\begin{figure*}[t]
    \centering
  \includegraphics[width=\linewidth]{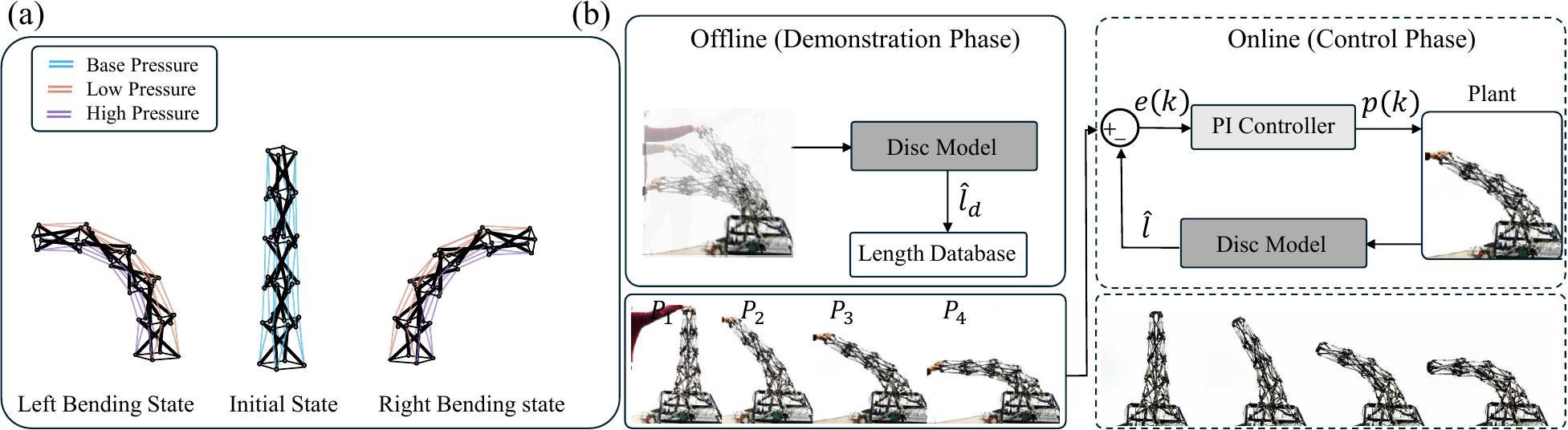}

    \caption{
    Bending behavior and control architecture of the Tensegrity manipulator. (a) The manipulator achieves bending through asymmetrical active cable lengths. (b) The left panel shows the demonstrated sequence of postures $P_1$-$P_4$ with the 40 active cable lengths recorded for each posture. The right panel shows the control scheme, which uses the demonstrated active cable lengths as a reference to achieve the desired postures through closed-loop control.
    }
    \label{fig:PIcontrol}
    \end{figure*}

\subsection{Shape Based Control}

 To control the posture of the tensegrity manipulator, we use the lengths of each active cable, where asymmetrical cable length produces directional bending of the structure, shown in Figure.~\ref{fig:PIcontrol} (a). Let $\hat{l} \in \mathbb{R}^{40}$ be the current lengths, where each element represents the Euclidean distance between a pair of joints located on two adjacent disks. A desired posture is represented by $\hat{l}_d \in \mathbb{R}^{40}$, which is obtained offline via kinesthetic teaching. For each demonstrated posture, the active cable lengths obtained from the estimator are stored as a reference. During the control phase, the same estimator is used to compute the current cable lengths, and a Proportional Integral (PI) controller regulates the pressures based on cable length errors, driving the manipulator to the desired postures.

\begin{figure}[h]
    \centering
  \includegraphics[width=\linewidth]{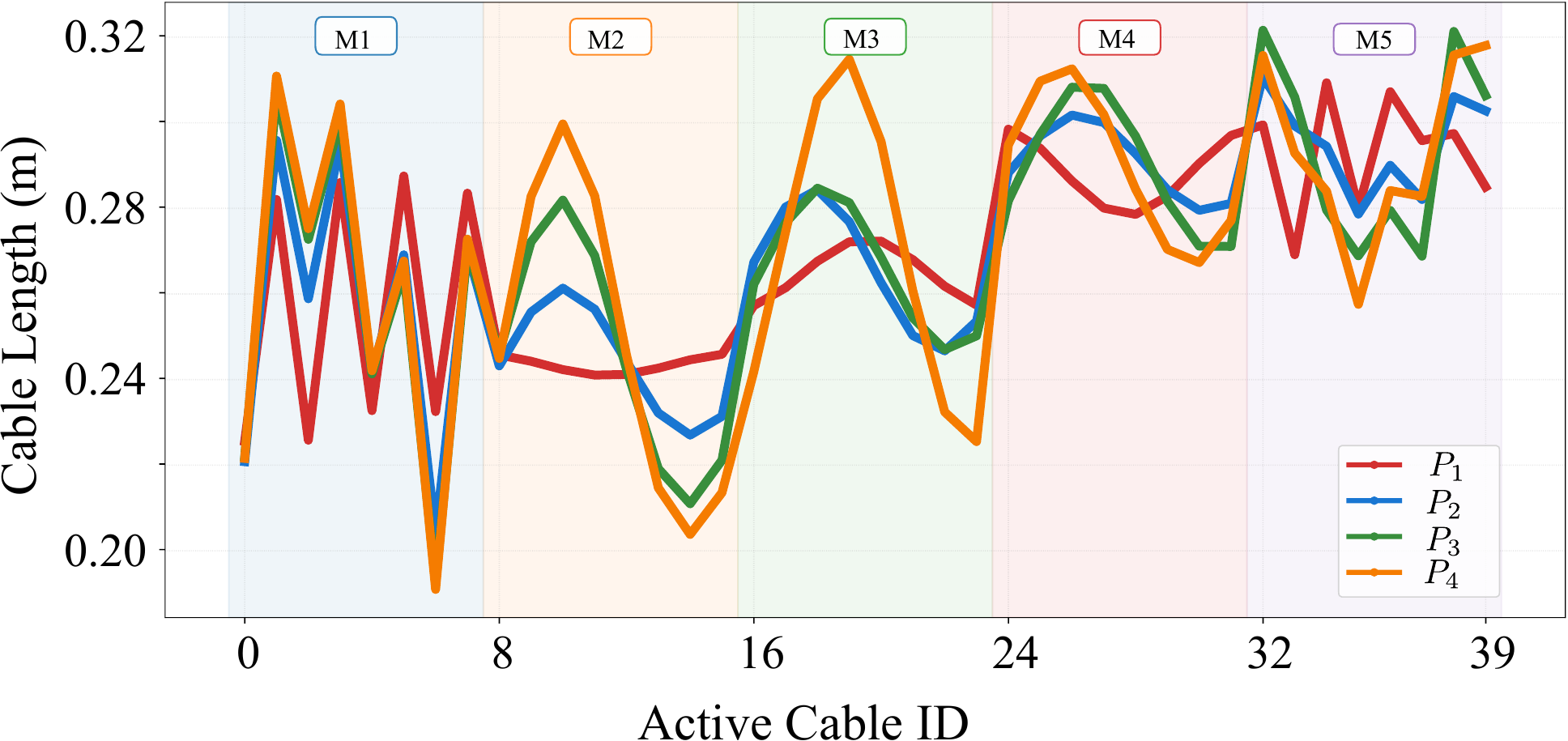}

    \caption{Desired active cable lengths corresponding to the desired postures $P_1$, $P_2$, $P_3$, and $P_4$. Each module (M1-M5) contains 8 active cables.}
    \label{fig:active_cable_lengths}
 \end{figure}
   \begin{table}[h]
\centering
\caption{Mean errors of active cables Across Postures}
\label{tab:active_cable_erros}
\begin{tabular}{lcc}
\toprule
{Posture ($P$)} & {Mean Error (mm)} & {Max Error (mm)} \\
\midrule
1 & 3.2 & 10.5 \\
2 & 4.6 & 12.0 \\
3 & 4.7 & 11.8 \\
4 & 5.0 & 11.4 \\
\bottomrule
\end{tabular}
\end{table}

 \subsubsection*{Proportional Integral (PI) based posture control}
The Proportional Integral (PI) controller is applied independently for each cylinder, and the control law is written as 
\[
p(k) = p(k-1) - K_p\, e(k) - K_i\, I(k)
\]
where $K_p$ and  $K_i $ are controller gains, $e(k)$ is the error, $e(k) = \hat{l}_d - \hat{l} $, and the integral $I(k)$ of error can be approximated in discrete time using the Euler method.
\[
I(k) = I(k-1) + e(k)\, T
\]
 Where $T$ is the sampling period. The PI controller achieves the sequence of desired postures in a quasistatic manner.

 Figure.~\ref{fig:active_cable_lengths} shows the desired active cable lengths obtained for the target postures. The left panel of Figure.~\ref{fig:PIcontrol} (b) shows the desired postures ($P_1$-$P_4$) while the right panel shows the resulting controlled postures corresponding to the desired postures. Table~\ref{tab:active_cable_erros} reports the errors in active cable lengths across the postures.

\section{Conclusion}
 We present an optimisation-based shape estimator for modular tensegrity robots that uses onboard IMU measurements and geometric constraints of the tensegrity structure. The proposed method is evaluated on both a simple single-layer class 1 tensegrity structure and a five-layer tensegrity continuum manipulator. For the modular tensegrity, the estimator successfully recovers the full tip-pose from arbitrary initial conditions, achieving a positional RMSE of 0.0913 m and orientation errors of 0.216 rad (tilt) and 0.205 rad (yaw). We also demonstrated posture control using a PI controller based on the estimated active cable lengths. In future work, we aim to compensate for the dynamic accelerations to maintain the estimation accuracy during high-speed motions.

\bibliographystyle{IEEEtran}
\bibliography{references} 

\end{document}